\documentclass[letterpaper, 10 pt, conference]{ieeeconf}
\IEEEoverridecommandlockouts                              
\usepackage{comment}
\usepackage{url}

\usepackage{multirow}
\usepackage{graphicx}
\usepackage{epsfig}
\usepackage{epic,eepic}
\usepackage{times}
\usepackage{mathptmx}
\usepackage{cite}
\usepackage{color}

\usepackage{times}

\definecolor{red}{rgb}{1,0,0}
\definecolor{green}{rgb}{0,1,0}
\definecolor{blue}{rgb}{0,0,1}
\definecolor{violet}{rgb}{1,0,1}
\definecolor{cyan}{cmyk}{1,0,0,0}
\definecolor{magenta}{cmyk}{0,1,0,0}
\definecolor{yellow}{cmyk}{0,0,1,0}

\definecolor{white}{rgb}{1,1,1}

\newcommand{\CO}[1]{}

\newcommand{\CommentOut}[1]{}

\newcommand{\noeditage}[1]{#1} \newcommand{\editage}[1]{}

\newcommand{\islarge}{}

\begin{document}

\newcommand{\titleauthor}[2]{\title{\bf\Large%
#1}%
\author{#2}%
\maketitle%
}

% \twocolumn

\newcommand{\FIG}[3]{
\begin{minipage}[b]{#1cm}
\begin{center}
\includegraphics[width=#1cm]{#2}\\
{\scriptsize #3}
\end{center}
\end{minipage}
}

\newcommand{\FIGU}[3]{
\begin{minipage}[b]{#1cm}
\begin{center}
\includegraphics[width=#1cm,angle=180]{#2}\\
{\scriptsize #3}
\end{center}
\end{minipage}
}

\newcommand{\FIGm}[3]{
\begin{minipage}[b]{#1cm}
\begin{center}
\includegraphics[width=#1cm]{#2}\\
{\scriptsize #3}
\end{center}
\end{minipage}
}

\newcommand{\FIGR}[3]{
\begin{minipage}[b]{#1cm}
\begin{center}
\includegraphics[angle=-90,width=#1cm]{#2}
\\
{\scriptsize #3}
\vspace*{1mm}
\end{center}
\end{minipage}
}

\newcommand{\FIGRt}[3]{
\begin{minipage}[t]{#1cm}
\begin{center}
\includegraphics[angle=-90,clip,width=#1cm]{#2}\vspace*{1mm}
\\
{\scriptsize #3}
\vspace*{1mm}
\end{center}
\end{minipage}
}

\newcommand{\FIGRm}[3]{
\begin{minipage}[b]{#1cm}
\begin{center}
\includegraphics[angle=-90,clip,width=#1cm]{#2}\vspace*{0mm}
\\
{\scriptsize #3}
\vspace*{1mm}
\end{center}
\end{minipage}
}

\newcommand{\FIGC}[5]{
\begin{minipage}[b]{#1cm}
\begin{center}
\includegraphics[width=#2cm,height=#3cm]{#4}~$\Longrightarrow$\vspace*{0mm}
\\
{\scriptsize #5}
\vspace*{8mm}
\end{center}
\end{minipage}
}

\newcommand{\FIGf}[3]{
\begin{minipage}[b]{#1cm}
\begin{center}
\fbox{\includegraphics[width=#1cm]{#2}}\vspace*{0.5mm}\\
{\scriptsize #3}
\end{center}
\end{minipage}
}

\islarge{
\LARGE
}

\titleauthor{
Training Self-localization Models for Unseen Unfamiliar Places \\
via Teacher-to-Student Data-Free Knowledge Transfer 
% Open-World Distributed Robot Localization via Recursive Data-Free Distillation
}{Kenta Tsukahara ~~~~~ Kanji Tanaka ~~~~Daiki Iwata \thanks{Our work has been supported in part by JSPS KAKENHI Grant-in-Aid for Scientific Research (C) 20K12008 and 23K11270.}\thanks{$*$%
K. Tsukahara, K. Tanaka, and D. Iwata are with Robotics Coarse, 
Department of Engineering, University of Fukui, Japan. 
{\tt\small{\{ha200973, tnkknj, ha200213\}@u-fukui.ac.jp}}}}

\newcommand{\figC}{
\begin{figure*}[t]
\hspace*{-5mm}
\begin{minipage}[b]{9.2cm}\begin{center}
\scriptsize
\FIGR{9}{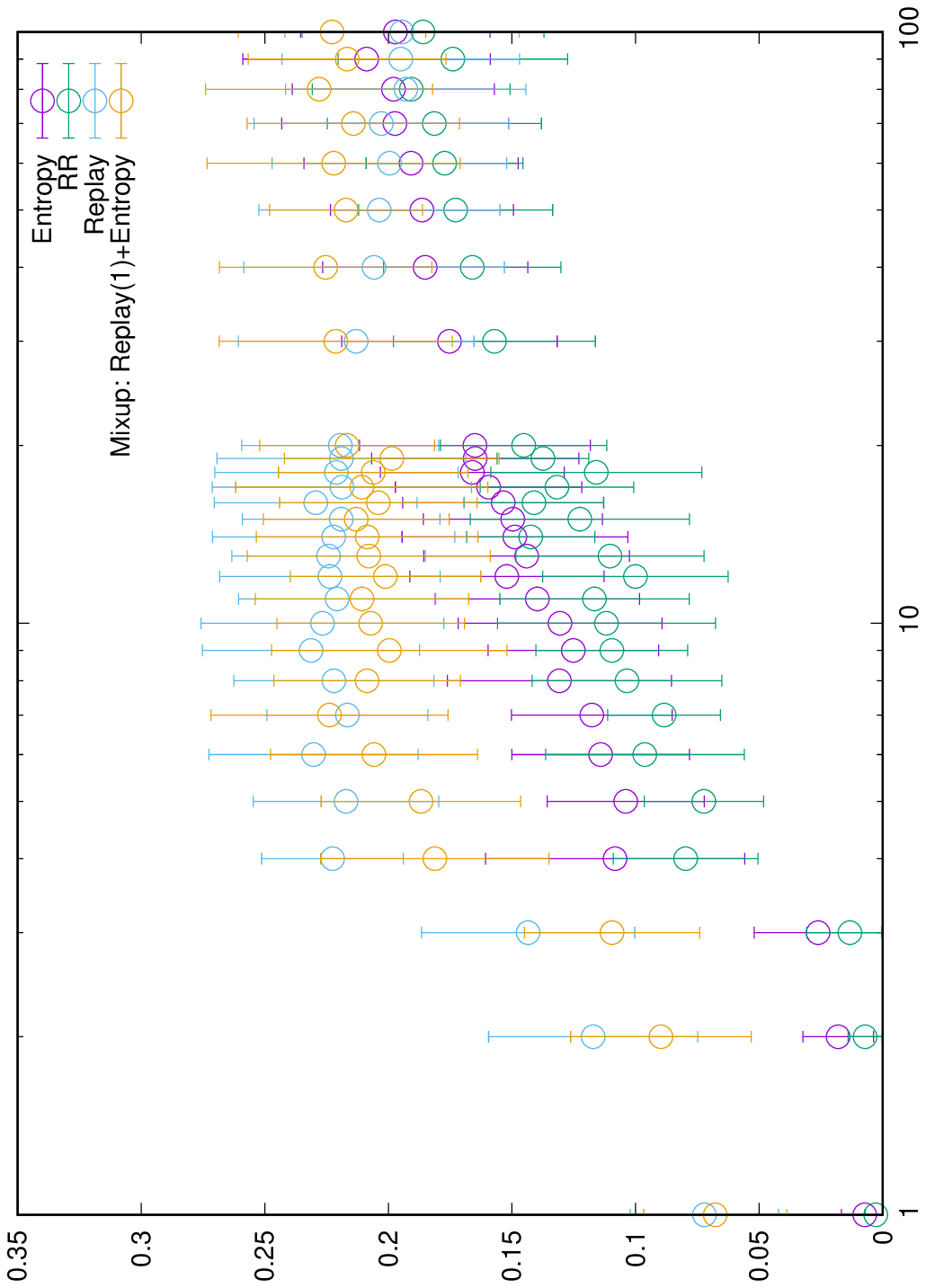}{}\\
(a) Performance results
\end{center}
\end{minipage}\hspace*{-5mm}%
\begin{minipage}[b]{9.5cm}
\begin{center}
\scriptsize
\FIGR{4.5}{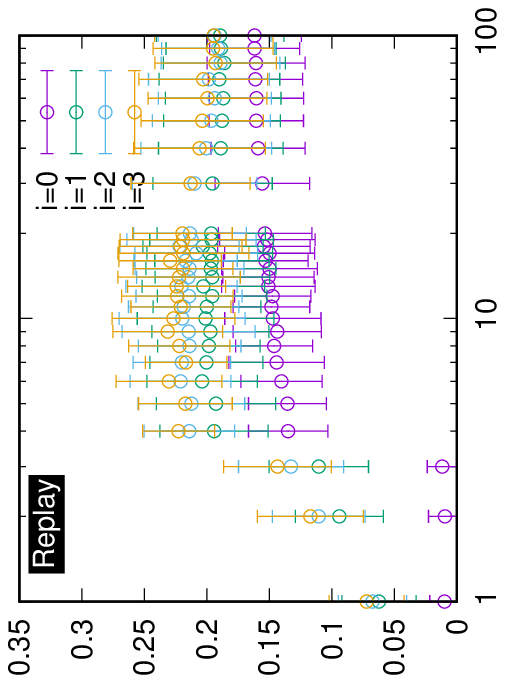}{}
\FIGR{4.5}{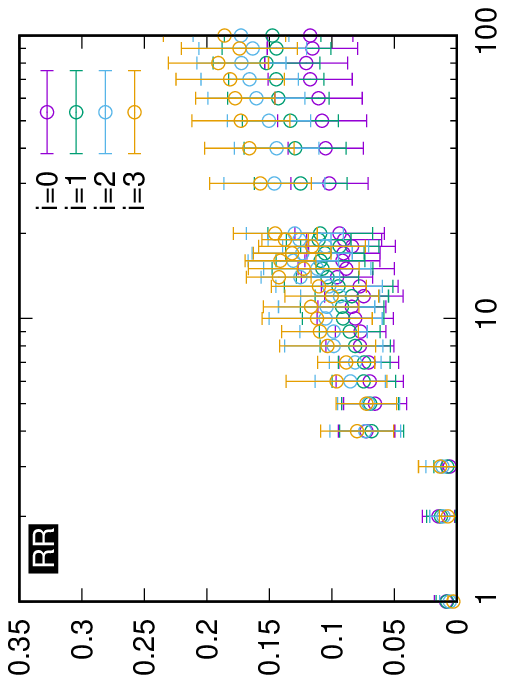}{}\\
\FIGR{4.5}{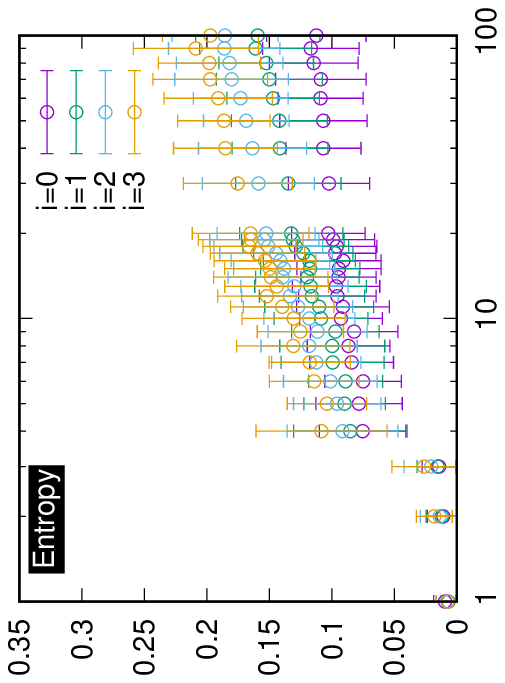}{}
\FIGR{4.5}{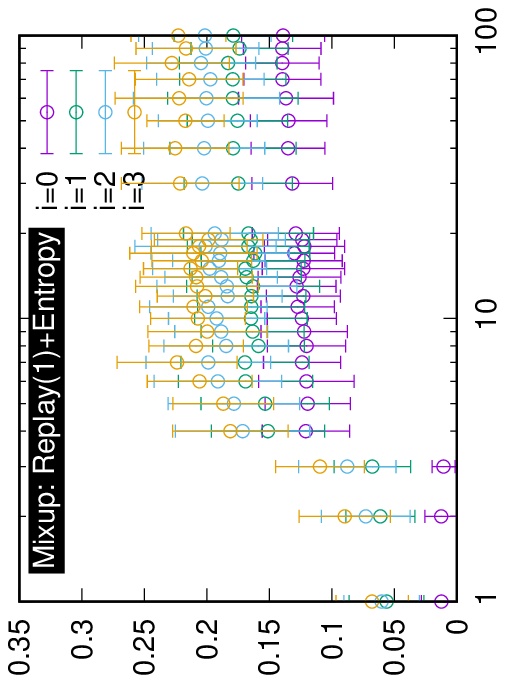}{}\\
(b) Effects of catastrophic forgetting
\end{center}
\end{minipage}
\caption{Top-1 accuracy performance vs. KT cost ($T$).}\label{fig:C}
\end{figure*}
}

\newcommand{\figB}{
\begin{figure}[t]
\begin{center}
\hspace*{5mm}\FIG{7}{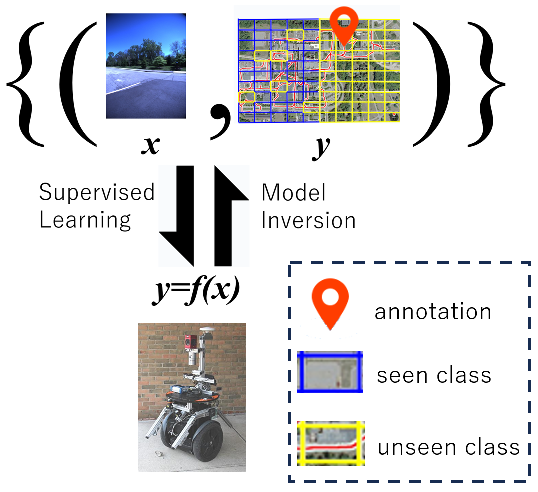}{}
\caption{%
Our goal is to design an excellent questioner (i.e., student) that can obtain a sequence of question $x_t$ and answer $y_t$ pairs via interactions with a blackbox teacher $y=f(x)$ such that the obtaiend samples $\{(x_t, y_t)\}_{t=1}^T$ can then be used for supervised learning or distillation of the student self-localization model. This is relevant to the model inversion problem, except that we are targeting generic self-localization models. 
}\label{fig:B}
\vspace*{-5mm}
\end{center}
\end{figure}
}

\newcommand{\figD}{
\begin{figure}[t]
\begin{center}
\FIG{9}{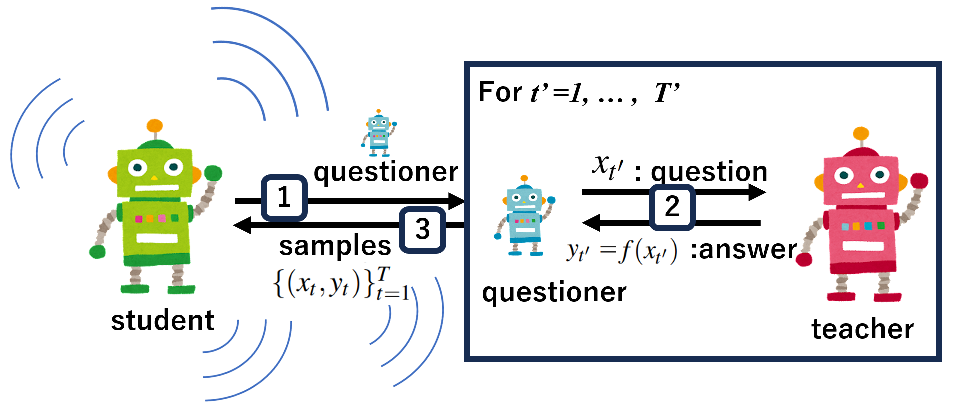}{}\vspace*{-3mm}\\
\caption{%
A typical communication protocol:
(1) The student robot sends the questioner (program code) to the teacher robot.
(2) The questioner repeats the question and answer session with the teacher.
(3) The student robot receives the selected sample via the KT channel.
In this work, we aim to obtain a small number of good samples.
}\label{fig:D}
\vspace*{-5mm}
\end{center}
\end{figure}
}

\begin{abstract}
A typical assumption in state-of-the-art self-localization models is that an annotated training dataset is available in the target workspace. However, this does not always hold when a robot travels in a general open-world. This study introduces a novel training scheme for open-world distributed robot systems. In our scheme, a robot (``student") can ask the other robots it meets at unfamiliar places (``teachers") for guidance. Specifically, a pseudo-training dataset is reconstructed from the teacher model and thereafter used for continual learning of the student model. Unlike typical knowledge transfer schemes, our scheme introduces only minimal assumptions on the teacher model, such that it can handle various types of open-set teachers, including uncooperative, untrainable (e.g., image retrieval engines), and blackbox teachers (i.e., data privacy). Rather than relying on the availability of private data of teachers as in existing methods, we propose to exploit an assumption that holds universally in self-localization tasks: ``The teacher model is a self-localization system" and to reuse the self-localization system of a teacher as a sole accessible communication channel. We particularly focus on designing an excellent student/questioner whose interactions with teachers can yield effective question-and-answer sequences that can be used as pseudo-training datasets for the student self-localization model. When applied to a generic recursive knowledge distillation scenario, our approach exhibited stable and consistent performance improvement. 
\end{abstract}

\section{Introduction}

Self-localization, that is, the problem of classifying a view image into predefined classes, is a fundamental problem in visual robot navigation and has important applications, including scene understanding, map building, and path planning. Most existing solutions, ranging from image retrieval engines \cite{ibowlcd} to ConvNet image classifiers \cite{planet}, aim to construct high-quality self-localization models using annotated training datasets for supervision. In a familiar workplace, this assumption actually holds, since pseudo-ground-truth viewpoints or place-labels can be reconstructed from a visual experience via structure-from-motion \cite{Brachmann_2021_ICCV}. Several state-of-the-art techniques can achieve excellent performances in such supervised settings, as discussed in \cite{itsc2018fang}. However, this is not the case in an unfamiliar workspace where no supervision is available. Therefore, the problem of self-localization remains largely unresolved.

In this study, teacher-to-student knowledge transfer (KT) or distillation (KD) for general open-world and distributed robot systems such as open-world distributed robot localization (OWDL) is considered as an alternative training setup. We observe that when humans travel around the open world, they often ask the people they meet in unfamiliar places for guidance. Therefore, we propose a similar knowledge-transfer scheme in which a student robot can consider other robots encountered in unfamiliar places as potential teachers and ask them to transfer knowledge about those places. Various types of teacher robots may exist. Some of them may be cooperative, whereas others may not. Some of them may be trainable (e.g., differentiable neural networks \cite{planet}), whereas others may not (e.g., image retrieval engines \cite{ibowlcd}). Some may have a known architecture, whereas others may have a blackbox architecture (i.e., data privacy). Therefore, we propose introducing minimal assumptions regarding potential teacher robots.

\noeditage{
\figB
}

This scenario is most relevant to the replay scheme \cite{ReplayCL} in the continual-learning (CL) field in which the model is supposed to be the primary source of knowledge to be transferred, and we wish to have no or minimal dependence on the availability of other training data or metadata. We are interested in CL, because there is always a risk of catastrophic forgetting \cite{catastrophic}, and because each time the student robot learns domain-/class-/vocabulary-specific knowledge from a new teacher, it risks forgetting the knowledge learned from previous teachers. The replay scheme can be applied to any architecture; therefore, we observe that it is more suitable for blackbox teachers than other typical CL methods, such as regularization \cite{RegularizationCL} or dynamic architecture \cite{DynamicArchitectureCL}. However, naive replay schemes assume the direct memorization of parts of the training sample set \cite{iros2015kanji} or the availability of teacher-specific sample generators \cite{generativeCL}, which significantly reduces the scope of applicability.

To address this issue, we focus on the problem of generating annotated training samples from a pre-trained teacher self-localization model (Fig. \ref{fig:B}). The problem of generating training samples from a given model (``Model Inversion" in Fig. \ref{fig:B}) can be considered as an inverse problem of supervised learning (``Supervised Learning" in Fig. \ref{fig:B}) \cite{DataFreeKnowledgeTransferSurvey31} and is related to recent studies on model inversion (MI) \cite{fredrikson2015model} and dataset reconstruction (DR) \cite{DatasetReconstruction}. However, these techniques focus on image generation and similar generative tasks and cannot be directly applied to self-localization tasks. Inter-robot communication channels (that could be used for KT) are usually limited \cite{li2023collaborative}, which prohibits the sending/reception of nonlightweight data, such as image data \cite{fredrikson2015model, DatasetReconstruction}. Furthermore, teacher robots can be of various types, as aforementioned. Rather than relying on the availability of private data of teachers as in existing MI/DR methods, we propose to exploit an assumption that holds universally in self-localization tasks: ``The teacher model is a self-localization system," and to reuse the self-localization system as a sole accessible teacher-side communication channel. 
Starting from a non-data-free scheme for benchmarking, we consider several KT schemes, those based on random walks in the input query space and those based on Entropy-based criteria, as well as a mixup of these schemes, and show that different schemes have different advantages and disadvantages. When applied to a generic recursive KD scenario \cite{itsc2019hiroki}, our approach exhibited stable and consistent performance improvement.

The main contributions of this work are:
(1)
We tackle the challenging problem of training a self-localization model in an unknown workspace where no annotated training dataset is available.
(2)
We introduce a practical teacher-to-student data-free knowledge transfer (DFKT) framework that introduces only a minimal assumption: ``The teacher model is a self-localization system,"  and reuses the teacher's available self-localization model as a sole accessible teacher-side communication channel;
(3)
We find that the proposed scheme consistently achieves stable performance improvements in a realistic recursive KD setup;

\section{Open-World Distributed Robot Localization (OWDL)}\label{sec:problem}

We begin with reviewing the well-known supervised learning (SL) setting of a single robot self-localization model that aims to classify an input scene 	(embedding) $x$ into one $y$ of predefined place-classes $C$ ($y\in{C}$), via supervised learning using a training set in the form of $\{(x, y)\}$. The experimental scenario follows the multi-teacher multi-student KT scenario using the NCLT dataset \cite{nclt}, derived from our previous work \cite{itsc2018fang}. The NCLT dataset contains long-term navigation data from a Segway robot with an onboard monocular front-facing camera navigating a university campus across 27 seasonal domains. The dataset can provide a pairing of the view image ($x$ in Fig. \ref{fig:B}) annotated with the ground-truth place class ($y$ in Fig. \ref{fig:B}) for each viewpoint on the robot's viewpoint trajectory for each of the 27 domains or sessions. The definition of the place-class set $C$ is assumed to be provided in advance and is consistent across robots. Notably, unlike common well-defined class definitions, such as country/region/postal codes, defining a place-class in a robot-centric coordinate system is itself an open problem in general \cite{iv2020kanji}. Moreover, students do not necessarily have access to meta-information such as the number of teachers, performance of individual teachers, number and ID of unseen unfamiliar place-classes, relative pose of teachers, and physical means of communication. Nevertheless, for simplicity, in this study, we do not explore this issue. We adopt simple grid-based place partitioning, in which the workspace is partitioned into a $10\times{10}$ grid of 100 place-classes in the bird's eye coordinate system.

We then describe a non-data-free KT \cite{ReplayCL} as a natural extension of the above setting. The basic idea is to maintain a portion of the training sample set used for SL as part of the model rather than discarding it after SL \cite{iros2015kanji}. Given such an augmented model, KT for retraining the student model when a new teacher is encountered is a simple procedure: merge the training samples that are included in the new teacher and previous student models, and distill them into a new student model.

We introduce a DFKT \cite{DataFreeKnowledgeTransferSurvey} as an extension of the above KT. The DFKT differs from the above KT only in that it generates (pseudo) training samples from the teacher and previous student models, via teacher-student-interactions (Section \ref{sec:DFKT}), rather than assuming that training samples are included in the teacher/student model. 

We specifically consider a typical communication protocol (Fig. \ref{fig:D}) in which a student sends a questioner (program code) $g$ to a teacher, and the teacher replies with samples $\{(x_t, y_t)\}_{t=1}^T$. Specifically, the questioner has $T'$ $({\ge}T)$ question-and-answer sessions with the teacher. At each question-and-answer step $t'$ ($\in[1,T']$), the questioner asks the next question $x_{t'}$ and the teacher returns the corresponding answers (e.g., $y_{t'}$). Except for the initial step $t'=1$, the questioner is allowed to access the question-and-answer history $(x_1, y_1)$ $\cdots$ $(x_{t'}, y_{t'})$. After the questioner-teacher-interaction is finished, the teacher sends to the student a collection of selected samples $\{(x_t, y_t)\}_{t=1}^T$, which is then used to train the student model via normal distillation process. Therefore, it is reasonable to measure the cost of KT based on the number $T$ $(\le{T'})$ of samples sent back from the teachers to the students.

Our goal is to improve the trade-off between self-localization performance (top-1 accuracy) and KT cost $T$ (teacher-to-student communication cost).

\noeditage{
\figD
}

\subsection{Self-localization Model}\label{sec:embed}

Although our framework is sufficiently general to be applied to generic blackbox teachers, to facilitate experimental investigation and analysis, we employ a specific visual embedding model recently developed in \cite{icte2022ohta}. The embedding model is based on scene graphs, which have proven their effectiveness in the field of visual self-localization \cite{X-view, E6}. It is trained as a scene graph classifier that maps the input scene to the class-specific probability map. The same graph convolutional neural (GCN) network architecture as in \cite{icte2022ohta} is used. A scene graph is a discriminative scene model that combines the advantages of pose-invariant local, condition-invariant global, and their hybrid part descriptors (i.e., the graph nodes), with the additional advantage of discriminatively describing the relationships (i.e., the graph edges) between these scene parts. Specifically, this model comprises (1) a scene graph generator that generates a scene graph from a scene and (2) a scene graph embedding that maps the scene graph to a class-specific probability map. 

The steps to scene graph generation are as follows: First, semantic labels are assigned to pixels using DeepLab v3+ \cite{deeplabv3plus2018}, pretrained on Cityscapes dataset. Then, regions smaller than 1,000 pixels (for 616$\times$808 image) are regarded as noise and removed. Subsequently, connected regions with the same semantic labels are identified  using a flood-fill algorithm \cite{DBLP:conf/cvpr/HeHZ19}, and each is assigned a unique region ID. Next, each region is connected to each of its neighboring regions by an edge. Finally, an image scene graph with image region nodes is obtained. Specifically, each node is described by a feature vector as follows. The semantic labels output by a semantic segmentation network \cite{deeplabv3plus2018} were re-categorized  into seven different semantic category IDs:  ``sky,"  ``tree,"  ``building,"  ``pole," ``road,"  ``traffic sign," and ``the others" which respectively correspond to the labels \{``sky"\}, \{``vegetation"\}, \{``building"\}, \{``pole"\}, \{``road," ``sidewalk"\}, \{``traffic-light," ``traffic-sign"\}, and \{``person," ``rider," ``car," ``truck," ``bus," ``train," ``motorcycle," ``bicycle," ``wall," ``fence," ``terrain"\} in the original label space. The location of the region center was quantized by a 3$\times$3 regular grid into nine ``bearing" category IDs. The region size was quantized into three ``size" category IDs:  ``large (larger than 150 K pixels),"  ``medium (50 K-150 K pixels),"  and ``small (smaller than 50 K pixels)". Finally, these semantic, bearing and size category IDs are combined to obtain a (7$\times$9$\times$3=) 189-dim 1-hot vector as the node descriptor. Ablation studies have proven that the proposed spatial features contribute to significant performance improvements. Details of this scene graph generation are described in \cite{irosw2022yoshida}.

The scene graph embedding is trained as follows. The graph convolution operation takes node $v_i$ in the graph and processes it in the following manner. First, it receives messages from nodes connected by the edge. The collected messages are then summed via the SUM function. The result is passed through a single-layer fully connected neural network followed by a nonlinear transformation for conversion into a new feature vector. In this study, we used the rectified linear unit (ReLU) operation as the nonlinear transformation. The process was applied to all the nodes in the graph in each iteration, yielding a new graph that had the same shape as the original graph but updated node features. The iterative process was repeated $L$ times, where $L$ represents the ID of the last GCN layer. After the graph node information obtained in this manner were averaged, the probability value vector of the prediction for the graph was obtained by applying the fully connected layer and the softmax function. For implementation, we used the deep graph library  \cite{wang2019dgl} on the Pytorch backend.

\section{Data-Free Knowledge Transfer Schemes (DFKT)}\label{sec:DFKT}

This section discusses schemes for effective interaction between the questioner (i.e., the student's substitute) with teachers, such that effective question-and-answer sequences are obtained for DFKT. As explained in \ref{sec:problem}, the goal is to achieve a good trade-off between KT cost and self-localization performance.

\subsection{Replay Scheme}\label{sec:replay}

First, we introduce the replay scheme \cite{ReplayCL}, a well-known baseline CL scheme that is useful for benchmarking but cannot deal with blackbox teachers because it is not data-free. The basic idea is to maintain a portion of the training sample set used to train the model as part of the model rather than discarding it after training \cite{iros2015kanji}. Such training samples serve as knowledge transferred from the teacher to the student and can naturally be used as a training sample for supervised learning or distillation. This approach is non-data-free and therefore has the best ability to avoid catastrophic forgetting among the approaches used in our experiments. On the other hand, an obvious limitation of this approach is the assumption that it assumes accessibility to the teacher's training samples.

\subsection{Reciprocal Rank (RR) Scheme}

The simplest possible data-free scheme is to use a random sample as a question $x$. In contrast to the expensive replay samples introduced in \ref{sec:replay}, such random samples do not rely on the a priori knowledge of the teacher model and are therefore data-free. However, we experimentally found that this simple random scheme performed poorly. Regardless of whether it is used alone or in combination with the replay scheme, the performance of the simple random scheme is so poor such that the entire framework fails.

Therefore, we introduce a reciprocal rank (RR) scheme as a simple extension that satisfies the data-free requirements. This approach is motivated by the fact that several visual embeddings, including those introduced in Section \ref{sec:embed}, serve as ranking functions. In the original study in which this embedded model was constructed \cite{icte2022ohta}, the output of the model was modeled as a class-specific RR vector, called reciprocal rank feature (RRF). Therefore, we found that sampling $x$ from the RRF space instead of from the entire input space yields samples with acceptable performance.

This RRF vector is already low-dimensional, but is well approximated by an even lower-dimensional $k$-hot RRF ($k$=10). Note that this $k$-hot RRF can be computed efficiently by performing selection algorithm on an $N$-dimensional noise vector. This compressed version of the RR scheme was experimentally proven to perform as well as the simple RR scheme described above. For more information on RRF, please see \cite{icte2022ohta}.

\subsection{Entropy Scheme}\label{sec:entropy}

Next, we introduce an Entropy-based scheme, where Entropy is calculated from the class-specific probability map $P(c|x)$ predicted for each class $c$ by the model for a certain input signal $x$: 
\begin{equation}
E(x) = -\sum_c {P(c|x)} ~{\log} \Big[ ~P(c|x)~ \Big], 
\end{equation}
and its inverse is used as a score of the likelihood that $x$ belongs to a seen class. In a recent study \cite{DBLP:journals/ral/KimPK19}, the Entropy measure is employed in self-localization for an alternative task of discovering unseen place-classes. In contrast, we propose using Entropy as a measure to determine whether the teacher model is familiar with an input signal. Specifically, the Entropy scheme differs from the RR scheme in the following ways: Instead of generating $T$ RR samples, a much larger number $T' (\gg{T})$ of RR samples $\{(x_t, y_t)\}_{t=1}^{T'}$ are generated and $T$ KT samples with highest scores are selected among the $T'$ samples.

The class imbalance problem is a serious issue in the implementation of this scheme. Naively sampled RR inputs often belong to the popular classes. Consequently, the sampling efficiency of unpopular classes is poor and may lead to severe class imbalance. To avoid this issue, we introduce undersampling of the samples of popular classes. 

Notably, the Entropy value calculation used by this scheme implicitly relies on an assumption that we have access to class-specific probability map $P(1|x_t)$$\cdots$$P(C|x_t)$ of teacher output. Note that while this assumption holds true for neural network type self-localization models, it does not necessarily hold for all self-localization models. For example, in bag-of-words image retrieval engine type teacher self-localization model (e.g., \cite{I5}), a class probability map may not be available, but 
only class-specific rank values or relevance scores, meaning that the RR scheme works but the Entropy scheme does not. In our future work, we would like to expand the Entropy scheme so that it can be applied to diverse types of teacher models.

\subsection{Mixup Scheme}

The disadvantage of replay schemes, that is, their prohibitively high sample maintenance costs, can be compensated for by combining them with other schemes that maintain samples more cheaply. Therefore, we introduce the mixup scheme as the fourth scheme, which combines the replay scheme with other schemes (RR or Entropy). Specifically, only a small number $R$ of samples per class (e.g., $R$=1) from the replay scheme is assumed to be available, and by mixing up those few replay samples with samples provided by other schemes, the required number of samples is generated.

\noeditage{
\figC
}

\section{Experiments}

We evaluated the proposed DFKT scheme (Section \ref{sec:DFKT}) experimentally using the generic application scenario called OWDL (Section \ref{sec:problem}).

\subsection{Settings}\label{sec:settings}

In this experiment, a recursive distillation scenario was assumed in which students and teachers were trained in a supervised manner and the student encountered up to three $i=1,2,3$ teachers sequentially. At the initialization stage $i=0$, students and teachers are trained using supervised learning in a subset of the place-classes of size$\le$100. This subset is referred to as the ``classes-in-charge" for students or teachers. By default, the number of the classes-in-charge of each student/teacher was ten, and the classes were randomly selected from among the 100 place-classes. Therefore, the classes-in-charge of the student and teachers may partially overlap. 

The sequential 27 sessions recorded over two years of NCLT were divided into one test session: ``2012/08/04," one visual embedding training session: ``2012/04/29" and 25 sessions for training student/teacher self-localization model: ``03/25," ``03/31," ``04/05," ``05/11," ``05/26," ``06/15," ``08/20," ``09/28," ``10/28," ``11/04," ``11/16," ``11/17," ``12/01," ``01/08," ``01/10," ``01/15," ``01/22," ``02/02," ``02/04," ``02/05," ``02/12," ``02/18," ``02/19," ``02/23," and ``03/17" (``MONTH/DAY" is used as a session's name for brevity), for which session IDs $s=0$, $\cdots$, 24 are respectively assigned. 

To investigate generalization performance on different training datasets, we considered six different sessions $s=0$, $\cdots$, 5 as the student training dataset. Accordingly, each $i$-th teacher's training session ($i\in[1,2,3]$) is determined as $\{(6i+s)$ mod $25\}$. Note that the student is trained by supervised learning only in the initialization stage $i$=0, and in subsequent stages $i=1,2,3$, the student is trained via DFKT from $i$-th teacher it encountered.

Among the class-specific samples generated from the independent test dataset (``2012/08/04"), those belonging to the union of classes-in-charge of students and teachers are used as the test set. Recalling that by dafault each teacher/student is responsible for 10 classes, the cardinality of the union is 40 if there is no overlap between classes. In such a case, a student at the initialization stage ($i=0$) who has not encountered any teacher yet has experienced only 10 classes out of 40, and can only expect top-1 accuracy performance of about 25\% at most.

For the mixup scheme, $R$ replay samples are included in the number $T$ of KT samples per class, and $R$ is set to 1 by default.

At each KT stage ($i=1,2,3$), the student robot encounters a new $i$-th teacher, then a two-step KT is performed. First, for each class in the new teacher's classes-in-charge, $T$ KT samples are generated using a scheme described in Section \ref{sec:DFKT}. Second, for each class that is not the new teacher's classes-in-charge and is the classes-in-charge of the previous version student denoted as $(i-1)$-th student, $T'$ RR samples are generated and added to the KT sample set. Note that for the second student-to-student KT step, we can assume no KT cost (communication cost) will be incurred, and thus the number $T'$ of such cost-free samples is set in large quantities ($T'$=100 samples per class).

To demonstrate that the self-localization performance does not depend on the particular GCN architecture in Section \ref{sec:embed}, we performed knowledge distillation \cite{hinton} from the GCN (in \ref{sec:embed}) to a multilayer perceptron (MLP) with 4,096-dimensional hidden layer, and the MLP was used as an alternative to GCN throughout the experiment. In preliminary experiments, it was proven that this MLP achieves performance comparable to GCN. The main difference between GCN and MLP is the architecture-specific internal signals, but this is not a serious issue since the DFKT schemes considered in this work does not rely on internal signals. A merit of using MLP in place of GCN is that signal analysis is easy and suitable for experimental evaluations. Since the training dataset generated by the scheme described above will suffer from class imbalance, a balanced pseudo-dataset is temporarily created by oversampling fewer classes and is used to train the MLP.

For KD from teacher MLP to student MLP, we again used the method of Hinton et al. \cite{hinton} because it is the best-known KD scheme. In this standard distillation scheme, the loss function consists of distillation loss and student loss, with soft labels for the former and pseudo labels for the latter as training signals. Note that the soft labels computation assumes the availability of the teacher's output layer probability map. To be independent of this assumption, the KD algorithm could be replaced with a more general recent KD algorithm, which is out of the scope of this study, and even if such replacement is performed, it is expected that the performance superiority and inferiority between the schemes will be maintained.

\subsection{Results}

First, we investigated the basic performance. Recall that the student encountered up to three $i=1,2,3$ teachers sequentially. The student at the initialization stage $i=0$ and each $i$-th teacher are trained with a rich annotated training set of the classes-in-charge (i.e., 100 samples per class). At each step $i=1,2,3$, the student encounters a new teacher and conducts a teacher-to-student KT using the questioner as a mediator. Details of KT and distillation follows the procedures in \ref{sec:problem} and \ref{sec:settings}. Four schemes were used to select query $x$: RR, replay, Entropy, and mixup. Figure \ref{fig:C} (a) shows the performance results.

The results show that all four schemes can achieve high performance when the number of samples per class increases. (1) Note that the replay scheme is not affected by catastrophic forgetting under a sufficiently large number of samples, so it can achieve performance equivalent to supervised learning. However, this scheme is not data-free and is therefore used in this work only as a technique for benchmarking. (2) Although the RR scheme is a data-free scheme and a quite simple scheme, it was able to achieve surprisingly good performance. This may be partly due to the use of a good quality visual embedding by the graph convolution scene graph classifier, leaving investigation into various embedding methods for future consideration. Another potential factor is that RRF was a good approximation of visual embedding. As also shown in the literature, RRF is a compact feature vector, and even random RR features could be a good approximation of visual embeddings at high frequency. (3) The Entropy scheme is a data-free scheme and has always marked the same or better performance than the RR scheme. In particular, when the number of samples per class is small, the performance is significantly higher than the RR scheme, and it is clear that this scheme has the ability to generate a small number of elite samples. As the number of samples per class increased, we were able to achieve performance approaching that of the replay scheme (non-data-free). However, as discussed in \ref{sec:entropy}, a potential concern with this scheme is that it relies on the availability of class-specific probability maps, which is a stronger assumption than the availability of class-specific rank values. (4) The mixup scheme can achieve versatility and good cost performance. It relies on strong assumptions about keeping training samples as part of the model, but is more practical in that it keeps the number of training samples constant. As expected, this scheme achieved performance approaching that of the replay scheme. However, since this scheme is not strictly data-free, its range of applications is narrower than that of the replay scheme or the RR scheme. In conclusion, the performance was highest in the order of replay scheme, mixup scheme, Entropy scheme, and RR scheme, but the versatility was in the opposite order.

We compared the ability to deal with catastrophic forgetting between different schemes. Catastrophic forgetting \cite{10377864}, where learning new knowledge causes one to forget what was previously learned, is a serious challenge in CL. In our case, we expect that a student model at the stage $i=3$ who has learned from all teachers will be most affected by catastrophic forgetting. A simple way to measure the magnitude of this effect is to compare the class-specific test performance between classes learned from the new teacher $i=3$ and classes learned from past teachers/student $i=0,1,2$. Figure \ref{fig:C} (b) shows the results. 

It can be seen that the performance of the replay scheme is hardly affected by the value of $i$, meaning that the performance is hardly affected by catastrophic forgetting. As expected, the remaining schemes performed worse than the replay scheme, and among them, the mixup scheme, Entropy scheme, and RR scheme had the best performance in order. It is also noteworthy that the mixup scheme had relatively little performance drop compared with the replay scheme. This is because superior samples from the replay scheme mitigated the effects of catastrophic forgetting.

\bibliographystyle{IEEEtran} 
\bibliography{reference}

\begin{thebibliography}{10}
\providecommand{\url}[1]{#1}
\csname url@rmstyle\endcsname
\providecommand{\newblock}{\relax}
\providecommand{\bibinfo}[2]{#2}
\providecommand\BIBentrySTDinterwordspacing{\spaceskip=0pt\relax}
\providecommand\BIBentryALTinterwordstretchfactor{4}
\providecommand\BIBentryALTinterwordspacing{\spaceskip=\fontdimen2\font plus
\BIBentryALTinterwordstretchfactor\fontdimen3\font minus
  \fontdimen4\font\relax}
\providecommand\BIBforeignlanguage[2]{{%
\expandafter\ifx\csname l@#1\endcsname\relax
\typeout{** WARNING: IEEEtran.bst: No hyphenation pattern has been}%
\typeout{** loaded for the language `#1'. Using the pattern for}%
\typeout{** the default language instead.}%
\else
\language=\csname l@#1\endcsname
\fi
#2}}

\bibitem{ibowlcd}
E.~Garcia{-}Fidalgo and A.~Ortiz, ``ibow-lcd: An appearance-based loop-closure
  detection approach using incremental bags of binary words,'' \emph{{IEEE}
  Robotics Autom. Lett.}, vol.~3, no.~4, pp. 3051--3057, 2018.

\bibitem{planet}
P.~H. Seo, T.~Weyand, J.~Sim, and B.~Han, ``Cplanet: Enhancing image
  geolocalization by combinatorial partitioning of maps,'' in \emph{Computer
  Vision - {ECCV} 2018 - 15th European Conference, Munich, Germany, September
  8-14, 2018, Proceedings, Part {X}}, ser. Lecture Notes in Computer Science,
  vol. 11214.\hskip 1em plus 0.5em minus 0.4em\relax Springer, 2018, pp.
  544--560.

\bibitem{Brachmann_2021_ICCV}
E.~Brachmann, M.~Humenberger, C.~Rother, and T.~Sattler, ``On the limits of
  pseudo ground truth in visual camera re-localisation,'' in \emph{Proceedings
  of the IEEE/CVF International Conference on Computer Vision (ICCV)}, October
  2021, pp. 6218--6228.

\bibitem{itsc2018fang}
N.~Yang, K.~Tanaka, Y.~Fang, X.~Fei, K.~Inagami, and Y.~Ishikawa, ``Long-term
  vehicle localization using compressed visual experiences,'' in \emph{21st
  International Conference on Intelligent Transportation Systems, {ITSC} 2018,
  Maui, HI, USA, November 4-7, 2018}.\hskip 1em plus 0.5em minus 0.4em\relax
  {IEEE}, 2018, pp. 2203--2208.

\bibitem{ReplayCL}
D.~Isele and A.~Cosgun, ``Selective experience replay for lifelong learning,''
  in \emph{Proceedings of the Thirty-Second {AAAI} Conference on Artificial
  Intelligence (AAAI-18)}.\hskip 1em plus 0.5em minus 0.4em\relax {AAAI} Press,
  2018, pp. 3302--3309.

\bibitem{catastrophic}
A.~Robins, ``Catastrophic forgetting in neural networks: the role of rehearsal
  mechanisms,'' in \emph{Proceedings 1993 The First New Zealand International
  Two-Stream Conference on Artificial Neural Networks and Expert Systems},
  1993, pp. 65--68.

\bibitem{RegularizationCL}
D.~L. Silver and R.~E. Mercer, ``The task rehearsal method of life-long
  learning: Overcoming impoverished data,'' in \emph{Advances in Artificial
  Intelligence, 15th Conference of the Canadian Society for Computational
  Studies of Intelligence, {AI} 2002, Calgary, Canada, May 27-29, 2002,
  Proceedings}, ser. Lecture Notes in Computer Science, vol. 2338.\hskip 1em
  plus 0.5em minus 0.4em\relax Springer, 2002, pp. 90--101.

\bibitem{DynamicArchitectureCL}
A.~Mallya and S.~Lazebnik, ``Packnet: Adding multiple tasks to a single network
  by iterative pruning,'' in \emph{2018 {IEEE} Conference on Computer Vision
  and Pattern Recognition, {CVPR} 2018, Salt Lake City, UT, USA, June 18-22,
  2018}.\hskip 1em plus 0.5em minus 0.4em\relax Computer Vision Foundation /
  {IEEE} Computer Society, 2018, pp. 7765--7773.

\bibitem{iros2015kanji}
K.~Tanaka, ``Cross-season place recognition using {NBNN} scene descriptor,'' in
  \emph{2015 {IEEE/RSJ} International Conference on Intelligent Robots and
  Systems, {IROS} 2015, Hamburg, Germany, September 28 - October 2,
  2015}.\hskip 1em plus 0.5em minus 0.4em\relax {IEEE}, 2015, pp. 729--735.

\bibitem{generativeCL}
H.~Shin, J.~K. Lee, J.~Kim, and J.~Kim, ``Continual learning with deep
  generative replay,'' in \emph{Advances in Neural Information Processing
  Systems 30: Annual Conference on Neural Information Processing Systems 2017,
  December 4-9, 2017, Long Beach, CA, {USA}}, 2017, pp. 2990--2999.

\bibitem{DataFreeKnowledgeTransferSurvey31}
G.~K. Nayak, K.~R. Mopuri, V.~Shaj, V.~B. Radhakrishnan, and A.~Chakraborty,
  ``Zero-shot knowledge distillation in deep networks,'' in \emph{Proceedings
  of the 36th International Conference on Machine Learning, {ICML} 2019, 9-15
  June 2019, Long Beach, California, {USA}}, ser. Proceedings of Machine
  Learning Research, vol.~97.\hskip 1em plus 0.5em minus 0.4em\relax {PMLR},
  2019, pp. 4743--4751.

\bibitem{fredrikson2015model}
M.~Fredrikson, S.~Jha, and T.~Ristenpart, ``Model inversion attacks that
  exploit confidence information and basic countermeasures,'' in
  \emph{Proceedings of the 22nd ACM SIGSAC conference on computer and
  communications security}, 2015, pp. 1322--1333.

\bibitem{DatasetReconstruction}
N.~Haim, G.~Vardi, G.~Yehudai, O.~Shamir, and M.~Irani, ``Reconstructing
  training data from trained neural networks,'' in \emph{Advances in Neural
  Information Processing Systems}, vol.~35.\hskip 1em plus 0.5em minus
  0.4em\relax Curran Associates, Inc., 2022, pp. 22\,911--22\,924.

\bibitem{li2023collaborative}
Y.~Li, Z.~Lyu, M.~Lu, C.~Chen, M.~Milford, and C.~Feng, ``Collaborative visual
  place recognition,'' 2023.

\bibitem{itsc2019hiroki}
T.~Hiroki and K.~Tanaka, ``Long-term knowledge distillation of visual place
  classifiers,'' in \emph{2019 {IEEE} Intelligent Transportation Systems
  Conference, {ITSC}}.\hskip 1em plus 0.5em minus 0.4em\relax {IEEE}, 2019, pp.
  541--546.

\bibitem{nclt}
N.~Carlevaris{-}Bianco, A.~K. Ushani, and R.~M. Eustice, ``University of
  michigan north campus long-term vision and lidar dataset,'' \emph{Int. J.
  Robotics Res.}, vol.~35, no.~9, pp. 1023--1035, 2016.

\bibitem{iv2020kanji}
K.~Tanaka, ``Self-supervised map-segmentation by mining minimal-map-segments,''
  in \emph{{IEEE} Intelligent Vehicles Symposium, {IV} 2020, Las Vegas, NV,
  USA, October 19 - November 13, 2020}.\hskip 1em plus 0.5em minus 0.4em\relax
  {IEEE}, 2020, pp. 637--644.

\bibitem{DataFreeKnowledgeTransferSurvey}
Y.~Liu, W.~Zhang, J.~Wang, and J.~Wang, ``Data-free knowledge transfer: {A}
  survey,'' \emph{CoRR}, vol. abs/2112.15278, 2021.

\bibitem{icte2022ohta}
T.~Ohta, K.~Tanaka, and R.~Yamamoto, ``Scene graph descriptors for visual place
  classification from noisy scene data,'' \emph{{ICT} Express}, vol.~9, no.~6,
  pp. 995--1000, 2023.

\bibitem{X-view}
A.~Gawel, C.~Del~Don, R.~Siegwart, J.~Nieto, and C.~Cadena, ``X-view:
  Graph-based semantic multi-view localization,'' \emph{IEEE Robotics and
  Automation Letters}, vol.~3, no.~3, pp. 1687--1694, 2018.

\bibitem{E6}
X.~Guo, J.~Hu, J.~Chen, F.~Deng, and T.~L. Lam, ``Semantic histogram based
  graph matching for real-time multi-robot global localization in large scale
  environment,'' \emph{{IEEE} Robotics Autom. Lett.}, vol.~6, no.~4, pp.
  8349--8356, 2021.

\bibitem{deeplabv3plus2018}
L.-C. Chen, Y.~Zhu, G.~Papandreou, F.~Schroff, and H.~Adam, ``Encoder-decoder
  with atrous separable convolution for semantic image segmentation,'' in
  \emph{ECCV}, 2018.

\bibitem{DBLP:conf/cvpr/HeHZ19}
Y.~He, T.~Hu, and D.~Zeng, ``Scan-flood fill(scaff): An efficient automatic
  precise region filling algorithm for complicated regions,'' in \emph{{IEEE}
  Conference on Computer Vision and Pattern Recognition Workshops, {CVPR}
  Workshops 2019, Long Beach, CA, USA, June 16-20, 2019}.\hskip 1em plus 0.5em
  minus 0.4em\relax Computer Vision Foundation / {IEEE}, 2019, pp. 761--769.

\bibitem{irosw2022yoshida}
M.~Yoshida, R.~Yamamoto, and K.~Tanaka, ``Highly compressive visual
  self-localization using sequential semantic scene graph and graph
  convolutional neural network,'' in \emph{13th IROS Workshop on Planning,
  Perception, Navigation for Intelligent Vehicle}, 2022.

\bibitem{wang2019dgl}
M.~Wang, D.~Zheng, Z.~Ye, Q.~Gan, M.~Li, X.~Song, J.~Zhou, C.~Ma, L.~Yu,
  Y.~Gai, T.~Xiao, T.~He, G.~Karypis, J.~Li, and Z.~Zhang, ``Deep graph
  library: A graph-centric, highly-performant package for graph neural
  networks,'' \emph{arXiv preprint arXiv:1909.01315}, 2019.

\bibitem{DBLP:journals/ral/KimPK19}
G.~Kim, B.~Park, and A.~Kim, ``1-day learning, 1-year localization: Long-term
  lidar localization using scan context image,'' \emph{{IEEE} Robotics Autom.
  Lett.}, vol.~4, no.~2, pp. 1948--1955, 2019.

\bibitem{I5}
M.~Cummins and P.~M. Newman, ``Appearance-only {SLAM} at large scale with
  {FAB-MAP} 2.0,'' \emph{Int. J. Robotics Res.}, vol.~30, no.~9, pp.
  1100--1123, 2011.

\bibitem{hinton}
G.~E. Hinton, O.~Vinyals, and J.~Dean, ``Distilling the knowledge in a neural
  network,'' \emph{CoRR}, vol. abs/1503.02531, 2015.

\bibitem{10377864}
M.~Kang, J.~Zhang, J.~Zhang, X.~Wang, Y.~Chen, Z.~Ma, and X.~Huang,
  ``Alleviating catastrophic forgetting of incremental object detection via
  within-class and between-class knowledge distillation,'' in \emph{2023
  IEEE/CVF International Conference on Computer Vision (ICCV)}, 2023, pp.
  18\,848--18\,858.

\end{thebibliography}

\end{document}